\documentclass[letterpaper, 10 pt, conference]{ieeeconf}  

\IEEEoverridecommandlockouts                              

\usepackage{graphicx} 
\usepackage{amsmath} 
\usepackage{amssymb}  
\usepackage{multirow}
\usepackage{gensymb}
\usepackage{kotex} 

\usepackage{color}
\definecolor{blue}{rgb}{0,0,0.6}
\definecolor{green}{rgb}{0,0.3,0}
\definecolor{red}{rgb}{0.6,0,0}
\definecolor{gray}{rgb}{0.4,0.4,0.4}
\definecolor{black}{rgb}{0,0,0}
\definecolor{lightgray}{rgb}{0.83, 0.83, 0.83}
\definecolor{purple}{rgb}{1,0,1}

\title{\LARGE \bf
Quadruped Locomotion on Non-Rigid Terrain using Reinforcement Learning
}

\author{Taehei Kim$^{1}$ and Sung-Hee Lee$^{1}$
\thanks{$^{1}$Graduate School of Cultural Technology, KAIST, Daejeon, South Korea.
        {\tt\small \{hayleyy321|sunghee.lee\}@kaist.ac.kr}}%
}

\begin{document}

\maketitle
\thispagestyle{empty}
\pagestyle{empty}

\begin{abstract}
Legged robots need to be capable of walking on diverse terrain conditions. In this paper, we present a novel reinforcement learning framework for learning locomotion on non-rigid  dynamic  terrains. 
Specifically, our framework can generate quadruped locomotion on flat elastic terrain that consists of a matrix of tiles moving up and down passively when pushed by the robot's feet. A trained robot with 55cm base length can walk on terrain that can sink up to 5cm. We propose a set of observation and reward terms that enable this locomotion; in which we found that it is crucial to include the end-effector history and end-effector velocity terms into observation. 
We show the effectiveness of our method by training the robot with various terrain conditions. 
\end{abstract}


\section{Introduction}

Quadruped robots have many advantages such as stability and terrain adaptivity, and thus researchers have put consistent efforts to enabling quadruped robot locomotion in various conditions and environments. 
Reinforcement learning is a powerful tool for this problem as it allows linear reward design, which reduces the burden of sophisticated design of controllers based on physics and finds control policies that are robust to the change of environments.
Recently, deep reinforcement learning (DRL)-based methods have shown significant advances in quadruped locomotion tasks, such as walking with faster speed \cite{Hwangboeaau5872}, recovery from falling \cite{lee2019robust}, manipulation \cite{shi2020circus} and walking on diverse terrains \cite{9028188} with faster convergence \cite{article,8967995}. 

\begin{figure}[t]
      \centering
      \vspace*{0.25cm}
      \framebox{\parbox{3.3in}{\includegraphics[width=1.0\linewidth]{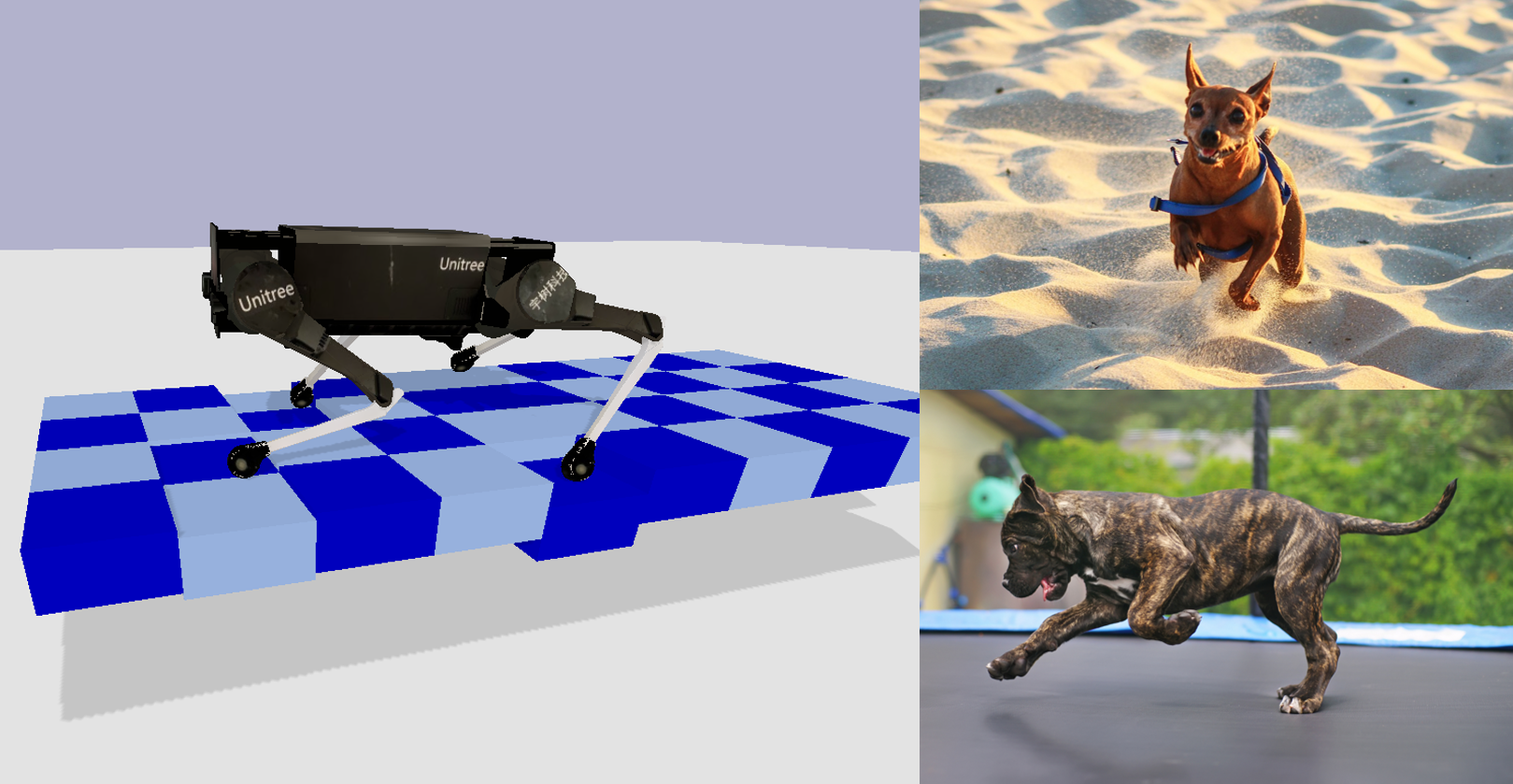}}}
      \caption{The real world includes various types of non-rigid terrain. We propose a DRL-based method for quadruped locomotion on flat elastic terrain. }
      \label{fig:teaser}
   \end{figure}

Our work explores to extend the range of environments for DRL-based quadruped locomotion: while the majority of previous studies assume rigid environments, we show the possibility of reinforcement learning frameworks of learning locomotion on non-rigid dynamic terrains. 
As our real-life environment includes grounds that plastically or elastically deform under pushing forces, such as sands, shaky bridges, or trampoline, the ability to locomote on non-rigid terrains is important for quadruped robots. To the best of our knowledge, our work is the first DRL framework, albeit on a simulated environment, that learns to walk on non-rigid terrains. Figure \ref{fig:teaser} shows a snapshot of our result where a virtual Laikago robot walks on the non-rigid terrain.

As our main contribution, we develop a framework that allows a quadruped robot to walk on a flat, non-rigid terrain, which is modeled as a tiled ground with each tile elastically sinking with varying stiffness. Specifically, we propose a set of observation and reward terms that enable the locomotion on the non-rigid terrain. Our experiments show that memorizing the history of end-effector positions with some time interval and the end-effector velocity are important. The effectiveness and limitation of our framework are examined by comparing results obtained for different set of terrain environments of training.


\section{Related Work}

Quadruped locomotion has been of great interest to researchers due to its possible utility. 
Starting from walking on flat rigid terrains, state-of-the-art methods aim to achieve more natural and agile movements in diverse environments. 
This section introduces studies that develop DRL-based methods for locomotion as well as studies to develop controllers for the similar environment like ours. 

\subsection{Walking on Rigid Flat Terrain}

Since Google DeepMind showed a DRL-based method for training virtual agents to locomote, DRL-based approaches to learning control policies for locomotion have been widely studied, e.g., \cite{heess2017emergence}. 
\cite{10.1145/2897824.2925881} trains virtual quadrupeds to run in 2D terrains with gaps, and \cite{10.1145/3072959.3073602} develops a two-level planning framework for a biped agent to walk. Biped locomotion is also studied in \cite{xie2018feedback} and \cite{Yu_2018} 

An important problem is to deploy a learned policy in virtual environment to real robots. 
One way to solve this is to train a controller with randomized settings such as including noise to secure versatility. \cite{DBLP:journals/corr/abs-1804-10332} successfully trains quadruped locomotion by including randomized physical environment and adding perturbations. \cite{inproceedings} randomizes the dynamics of the simulator to train a control policy for a fetching robot. 
Studies, such as  \cite{DBLP:conf/rss/HaarnojaHZTTL19} and \cite{ha2020learning}, show successful locomotion of small physical robot of Minitaur. \cite{yang2019data}  adopts the model-based reinforcement learning for Minitaur locomotion to reduce data collection during training for faster learning. \cite{ha2020learning} focuses on minimizing human interference when training on a real physical robot. \cite{Hwangboeaau5872} transfers policy learned in simulation to a physical robot using an actuator modeling. \cite{article} introduces a constrained policy optimization for faster convergence.  \cite{peng2020learning} applies animal behavior to real robots using the imitation learning approach, which uses reference motion data for an agent to follow to achieve challenging tasks.

\subsection{Walking on Various Terrain Environments}

Recent advancements in DRL-based methods advance further to the locomotion in diverse environments. In simulated environments, \cite{xie2020allsteps} introduces locomotion of biped agents on stairs with diverse heights and tilts. For quadruped agents, \cite{10.1145/3386569.3392433} introduces walking through obstacle-filled or slippery environments by using a mixture of imitation learning, general adversarial network and reinforcement learning in a simulated environment. \cite{9028188} introduces a method that mixes the benefit of the model-based planning and control approach and reinforcement learning to tackle environment with varying heights and gaps. 
\cite{rahme2020dynamics} introduces methods based on modulated Bezier curve gaits which enables uneven terrain locomotion using only inertial measures in physical robot. The simulation environment added the nominal clearance height, virtual ground penetration depth of the Bezier curve and residual foot displacements to the open-loop Bezier curve which might not be necessarily match with the real physical parameter. \cite{Leeeabc5986} develops a teacher-student reinforcement learning framework that can create foot trajectory traversing multiple environments including water, mud, and rock-filled terrains. \cite{shi2020circus} goes one step further by introducing a physical quadruped robot that can juggle a rubber ball. 

\subsection{Walking on Non-Rigid Terrain using Dynamic control}
Some studies tackle non-rigid terrain locomotion problem by developing new controllers or an efficient contact dynamics model. \cite{yang2020dynamic} shows a quadruped balancing on balls based on a model predictive control in a simulated environment. \cite{7139806} adopts a momentum-based controller to balance on non-rigid terrain using a relatively simple four-link planar robot. \cite{inproceedingsBosworth} introduces a controller tuned differently depending on rigid or non-rigid ground. \cite{Neunert_2018} relaxes hard constraints of contact dynamics so that their nonlinear model predictive control can be solved efficiently subject to contact, allowing non-rigid terrain locomotion. \cite{articleFahmi} introduces an online method by feeding terrain knowledge to a whole-body control for contact consistent solution. \cite{Chatzinikolaidis_2020} focuses on capturing the contact properties by developing a contact model that can be applied to direct trajectory planning. Our method tackles a similar environment where the terrain is elastically moving up and down when a quadruped steps on the ground. We propose another direction for the non-rigid terrain locomotion.

\section{Terrain and Robot Models}

   \begin{figure}[t]
      \centering
      \vspace*{0.25cm}
      \framebox{\parbox{3in}{\includegraphics[width=1.0\linewidth]{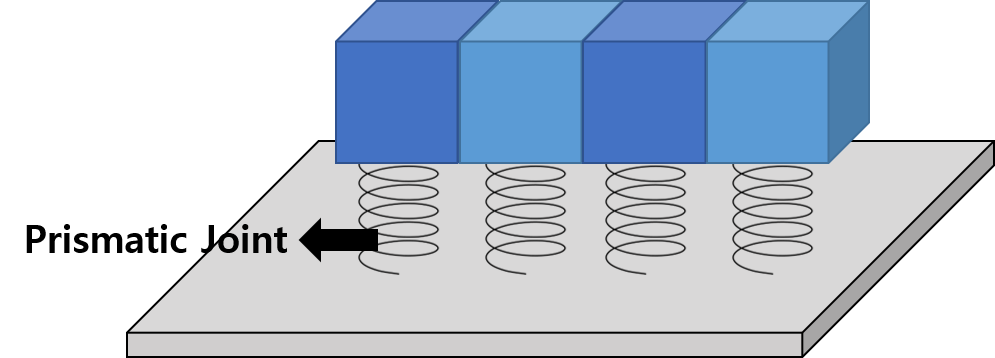}}}
      \caption{Our model for the flat elastic terrain. Each tile is connected to the floor by a spring-loaded prismatic joint.}
      \label{fig:terrain}
   \end{figure}

We construct a simulation environment by using PyBullet physics engine \cite{coumans2019} and employ a Laikago robot model \cite{chen_2018} as our target robot.
   
\subsubsection*{Construction of Non-Rigid Terrain}
The non-rigid terrain model that we design is a flat elastic terrain consisting of a matrix of tiles that can move up and down passively when a robot's foot pushes the tile, as shown in Fig.~\ref{fig:terrain}.
Each tile, of which width and length being 20 cm each, is connected to the flat rigid base via a prismatic joint with a spring. The stiffness of the spring is adjusted to control the amount of sinking. For instance, a terrain with a 5 cm sinking depth is created by setting the spring stiffness so that the average sinking depth of four tiles pressed by legs reaches 5 cm when the robot stands still. 

\begin{figure}[t]
      \centering
      \framebox{\parbox{3in}{\includegraphics[width=1\linewidth]{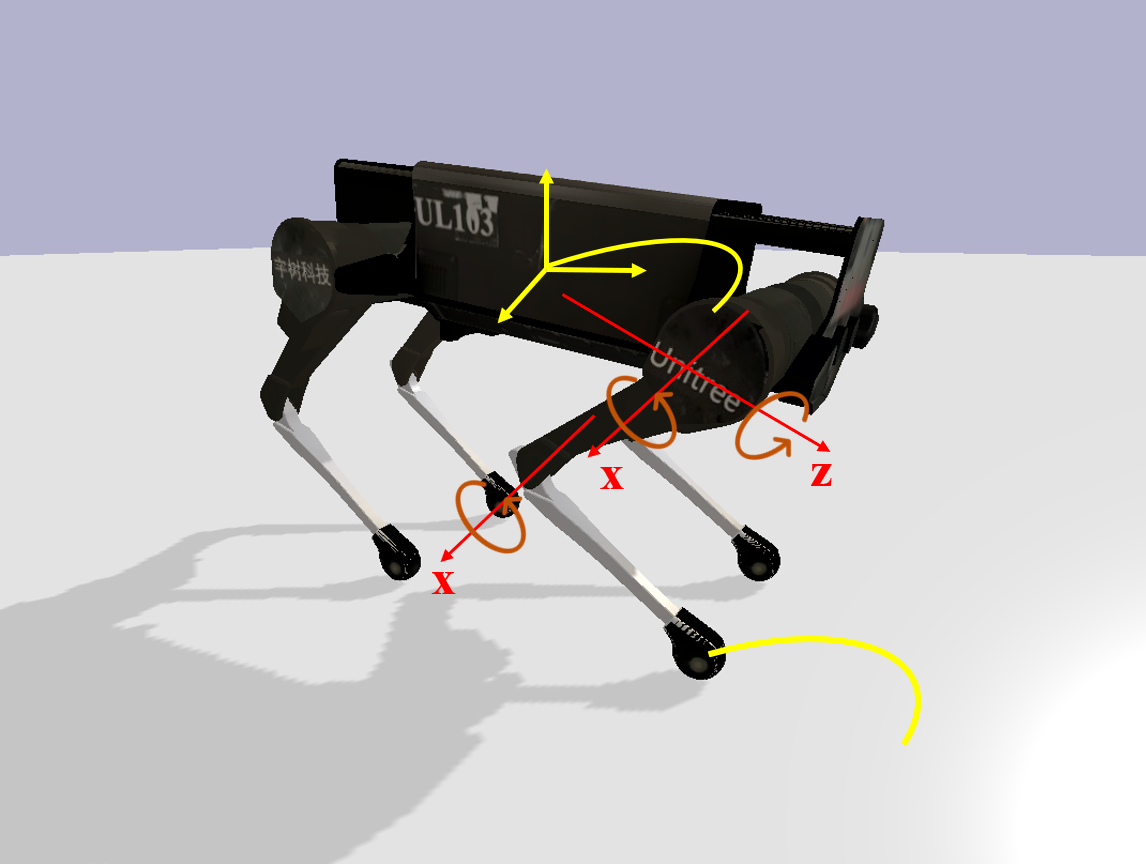}}}
      \caption{This figure shows the joint structure of the robot model. Yellow curves denote the trajectories of the base and the foot determined by the learned policy.}
      \label{fig:robot}
\end{figure}

\subsubsection*{Robot Model}
The Laikago robot model is about 25 kg and has 12 degrees of freedom (DoFs), with a 2-DoF shoulder and a 1-DoF knee joint in each leg. Figure~\ref{fig:robot} shows the joint structure of the robot. The length of the base is 55 cm and that of the leg is about 50 cm. 

\section{Training}

We now describe the design of our reinforcement learning framework for the quadruped on non-rigid terrains. 
A single locomotion cycle includes four phases: In each phase (0.75 sec), one foot takes off and lands while all other feet maintain contact with the ground. 
One action signal output by a control policy defines the movement of a robot for one entire phase, thus four consecutive action signals constitute one locomotion cycle.

\subsection{Action}

The target trajectories of the base and the moving foot in a phase are modeled with cubic Bezier curves, and our action defines the control points of the Bezier curves.
One advantage of the Bezier curve is that the resulting trajectories are smooth.
Cubic Bezier curve is chosen because it is the lowest order that allows enough diversity in the curve configuration.
A total of three 3D cubic Bezier curves are used for the base position, the base orientation, and the swing foot. In each Bezier curve, the first control point is automatically determined as the current value of the trajectory, and the remaining three control points are defined by the action. The coordinates of control points are expressed with respect to the base frame. As a result, one action includes a 27D vector.

\subsubsection*{Confining Action Space}
We found that providing a exploration range in the action space is important to obtain successful convergence of a learned policy.
In our case, the bounds describe the possible range of the Bezier curve control points.
Specifically, for the base position, the second and third control points are confined within [-6, 6] cm, and the last control point is limited to [-4, 8] cm in each coordinate axis from the current position, which is the coordinates of the first control point. For the base orientation, all three control points are bounded to [-0.3, 0.3] rad in each Euler angle coordinate from the current orientation. 
Confining action space for the foot trajectory is a bit more complicated. The height of the control points are bounded to [-15, 15] cm from the current height. On the other hand, the frontal and lateral coordinates are bounded with respect to their default positions when the robot takes the initial squatting pose. The lateral coordinate is bounded to [-15, 15] cm from the default position. For the frontal coordinate, the frontal feet is bounded to [-15, 15] cm while the hind feet is bounded to [$\delta$-15, $\delta$+15] cm from the default position. The amount of shift $\delta$ is set -2cm.


\subsubsection*{Motion Generation}
As a control signal, the desired angle for each joint is generated approximately every 4 milliseconds. For this, the created Bezier curves are divided into 180 points, and inverse kinematics is solved to obtain the desired joint angles to achieve the target configurations of the base and feet, specified by the Bezier curve points. The calculated joint angles are provided to the joint position controller in pyBullet to generate joint torques to achieve the desired joint angles. 

\subsection{Observation}

Our method generates action signals at relatively long time intervals of 0.75 seconds. The amount of  information that the robot can collect during the decision interval is very large. For example, robot can track its joint angles in every few millisecond. As such, we need to select only partial information as observation to keep the size of neural networks to a reasonable level.
The attributes in our observation space consist of only those measurable by the real robot. Our observed attributes can be categorized into three types; 1) values that are directly sensed by the robot 2) values that are induced from the sensed values 3) values related to the designed goals. Our observation consists of the following attributes which add up to a 102D vector. 

\begin{itemize}

\item The height of the base $h_b$, represented as the height of the four joints positions at the base from the terrain right below each joint (4D).
\item The orientation of the base represented by the direction of the gravity vector with respect to the base frame (3D).
\item The linear and angular velocities of the base (6D).
\item The pitch angle of the base (1D).
\item The history of the four end-effector positions in its base frame at the start of three previous phases and at the current time step (48D).
\item The history of the four end-effector positions in its base frame at 4 and 8 milliseconds before the current time step (24D).
\item The velocity of the four end-effectors (12D).
\item The direction of the goal (azimuth angle) $\phi_{g}$ from the base frame (1D).
\item The position of the goal $\rho_{g}$ in the base frame. Only the frontal and lateral coordinates are used (2D). 
\item Current phase (1D).

\end{itemize}

\subsection{Reward Function}
Our goal is to make the robot walk on non-rigid terrains without falling only with a small number of reward terms. A total of five reward terms are used. 

\subsubsection{Goal distance reward}
It checks whether the robot is moving towards the goal. 
\begin{equation}
R_d = \alpha_d \left( ||\rho_{g}|| - ||\rho_{g,p}||  \right) 
\end{equation}
where $\rho_{g,p}$ is the target position with respect to the robot's base frame at the start of the previous phase, and the scaling factor $\alpha_d=10$ if $||\rho_{g,t-1}|| > ||\rho_{g,t}||$, and $\alpha_d=1$, otherwise. 

\subsubsection{Goal orientation reward}
It checks whether the agent is heading in the right direction by giving a positive reward if the azimuth angle of the goal from the base is less than 10 degrees.
\begin{equation}
R_o =\max \left ( 0, 0.02\times(10 - |\phi_{g}|) \right )
\end{equation}

\subsubsection{Minimum height reward}
It promotes that the robot's base is above the minimum height from the terrain. If the height $h_b$ of the base is larger than 25 cm, the robot receives $R_s = 0.1$.

\subsubsection{Torque minimizing reward}
It encourages the robot to use less torque while achieving the goal. 
\begin{equation}
R_t = \max \left ( 0, 0.004\times\left ( \tau_{thresh} - \tau_{ave} \right ) \right ), 
\end{equation}
where $\tau_{ave}$ is the average magnitude of the joint torque vector during a phase and  $\tau_{thresh}=140$ in our experiment.

\subsubsection{Roll angle reward}
It encourages the robot to stabilize its roll angle. 
\begin{equation}
R_r =\max \left ( 0, 2\times(0.1 - |\varphi|) \right )
\end{equation}
where $\varphi$ is the roll angle represented in radian angle.

The total reward is thus set as $R=R_d+R_o+R_s+R_t+R_r$.

\subsection{Termination Condition}
We employ early termination to avoid falling into local minima and to enhance the sample efficiency, as suggested by \cite{10.1145/3197517.3201311}, \cite{10.1145/3355089.3356499}, and \cite{10.1145/3386569.3392381}. 
In our framework, if one of the following conditions is met, the current episode is terminated with the agent receiving -10 and the training is restarted from a new initial state. Thus, the policy is trained not to fall into early termination conditions.

\begin{itemize}
    \item The base height decreases below 20 cm, which indicates that the robot's configuration is near to the joint limit.
    \item The pitch angle of the base exceeds an allowed range (-15 to 15 degrees), which indicates the robot is inclined too much.
    \item Any link except end-effector collides with the ground. 
\end{itemize}

\subsection{Training and Implementation Details}

We first initialize our robot to a stable squat pose with a low center of mass height, as shown in Fig.~\ref{fig:robot}. 
The robot has a fixed phase transition order: it moves the legs in the order of front-left, rear-right, front-right, and rear-left.

In the training stage, we use 4 types of non-rigid terrains with varying sinking depths of 2, 3, 4 and 5 cm as well as a rigid terrain. 
The training always starts with a terrain with 2 cm sinking depth but after that the terrain type is changed randomly every N (2 or 8) meters. The initial position of the robot is randomly changed as well.  
We train our robot by giving a specific target in 2 meters front. 
Once it reaches its initial target, the next goal is given again in 2 meters ahead repeatedly.

We use OpenAI Gym \cite{1606.01540} to create the learning environment and adopt PPO algorithm \cite{schulman2017proximal} provided by Stable Baselines \cite{stable-baselines}. The policy and value networks have an identical structure of 2 hidden layers with 256 and 128 perceptrons each. We use tanh for the activation function. The discount factor is 0.95 and the policy learning rate is $2 \times 10^{-4}$. The size of minibatch is 4096 and the PPO epoch is 10.
\section{Results}

   \begin{figure}[t]
      \centering
      \vspace*{0.25cm}
      \framebox{\parbox{3.2in}{\includegraphics[width=1.0\linewidth]{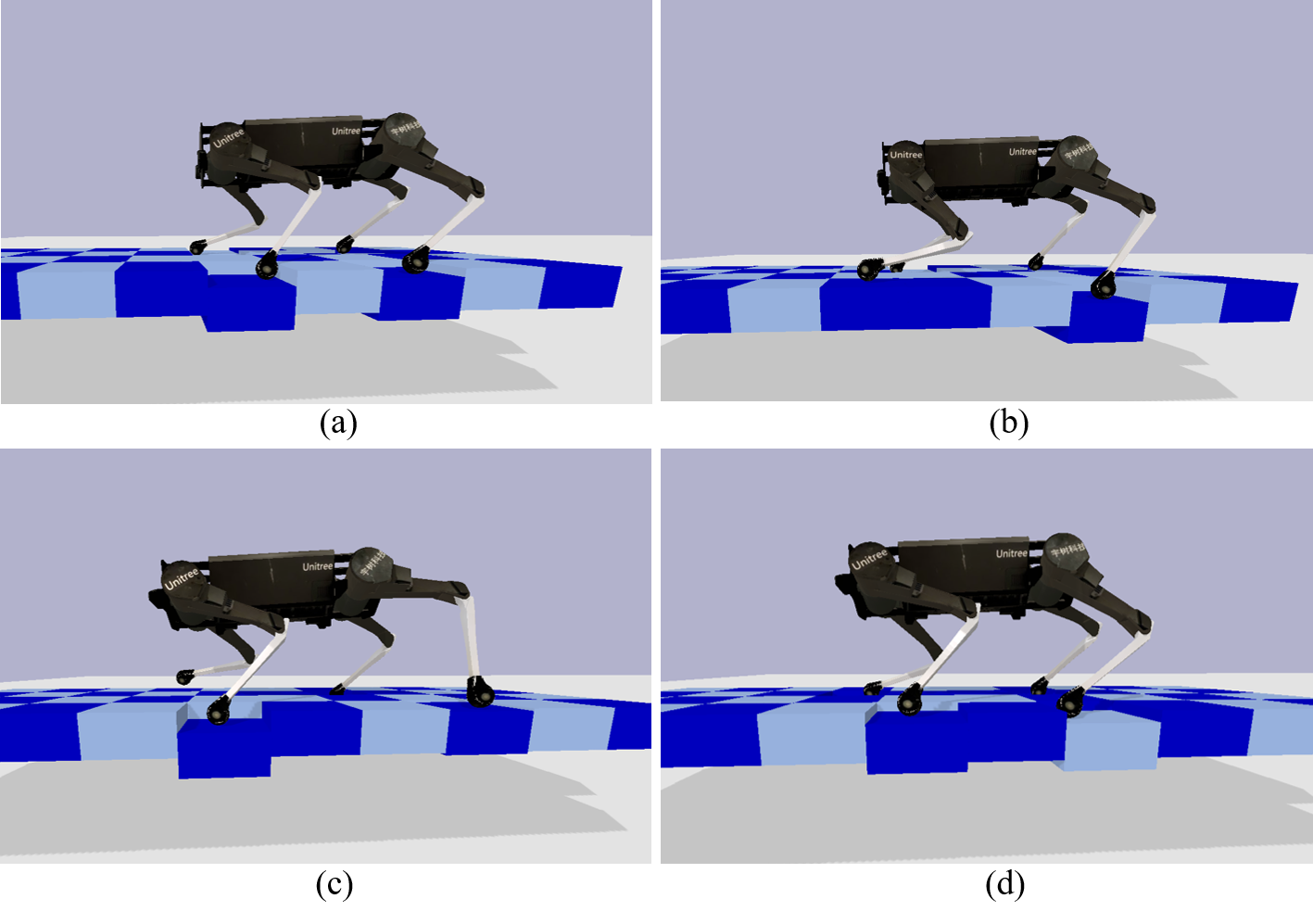}}}
      \caption{From (a) to (d), walking on non-rigid terrain with a 5 cm sinking depth.}
      \label{fig:5cmTerrain}
   \end{figure}

   \begin{figure}[t]
      \centering
      \vspace*{0.25cm}
      \framebox{\parbox{3.2in}{\includegraphics[width=1.0\linewidth]{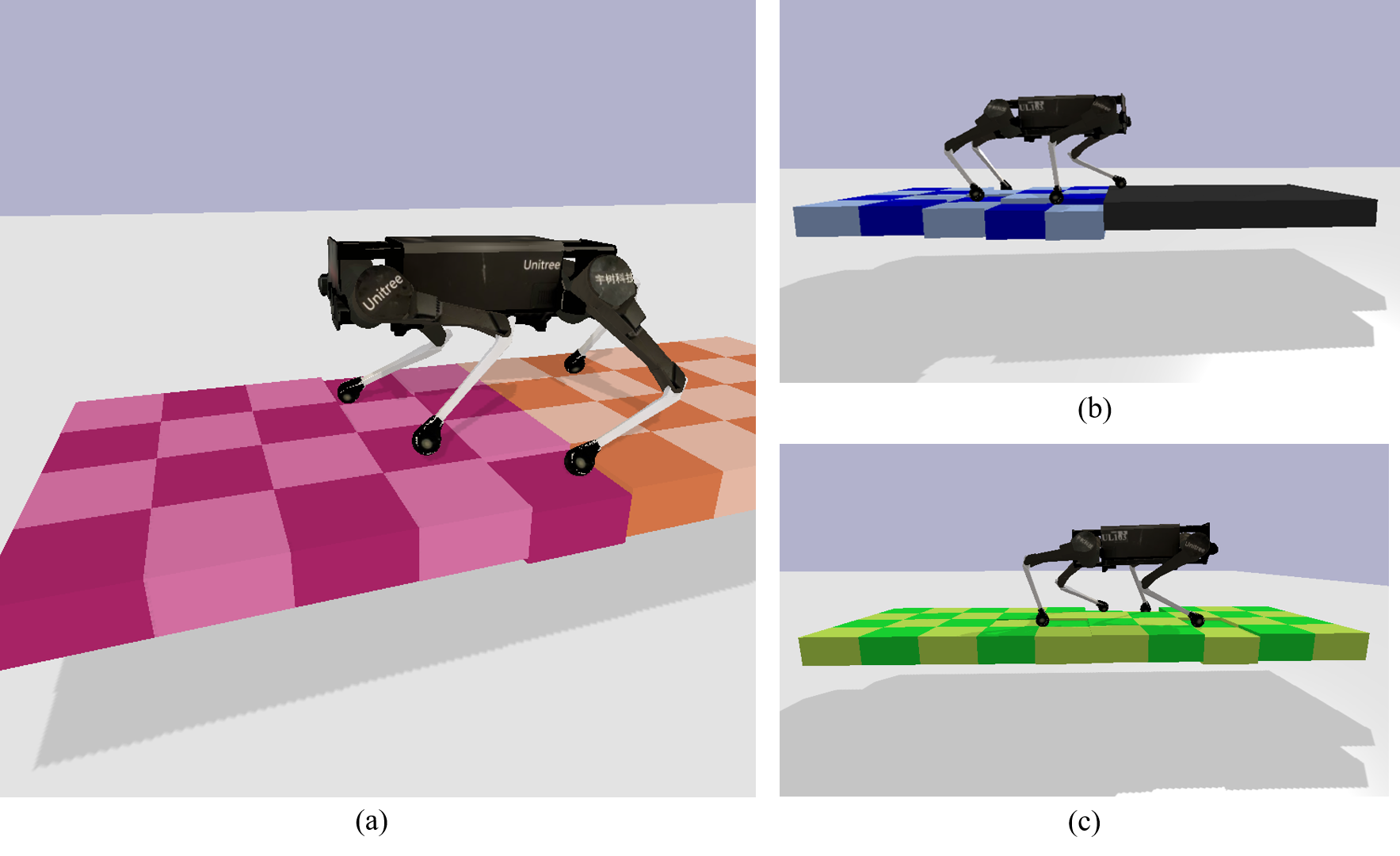}}}
      \caption{Walking on non-rigid terrain where the sinking depth changes every two meters. (a) Transition from 3 cm to 2 cm sinking depth. (b) Transition from 5 cm to rigid ground. (c) Walking on 4 cm sinking depth terrain.}
      \label{fig:multipleTerrain}
   \end{figure}

In this section, we examine the characteristics and the effectiveness of our method. For graphical view, Fig. \ref{fig:5cmTerrain} shows our result of quadruped walking on sinking depths of 5 cm, and Fig. \ref{fig:multipleTerrain} shows quadruped walking on terrain with varying stiffness.
We first examine the characteristics of our framework by analysing the trajectories of the base height and the target landing height of each foot on different terrains. Second, we identify some crucial components in our method. Third, we discuss the effect of other observation parameters that improve the stability of locomotion. Lastly, we share our experiments of increasing the observation space.

\subsection{Training on Different Terrains}

We examine how our method results in different policies depending on the terrain conditions for training.
Our training environments include four scenarios. 
\begin{itemize}
    \item $T_{v2}$ (Baseline): Terrain whose stiffness changes every 2 meters.
    \item $T_{v8}$: Terrain whose stiffness changes every 8 meters.
    \item $T_c^2$: Terrain of constant stiffness with 2 cm sinking depth.
    \item $T_c^5$: Terrain of constant stiffness with 5 cm sinking depth.
\end{itemize}
For $T_c^5$, training succeeds with curriculum learning by first training with $T_{v2}$, followed by training with $T_c^5$. To compare side by side, we also apply the same curriculum learning for $T_c^2$.

\begin{figure}[t]
    \centering
    \includegraphics[width=1\linewidth]{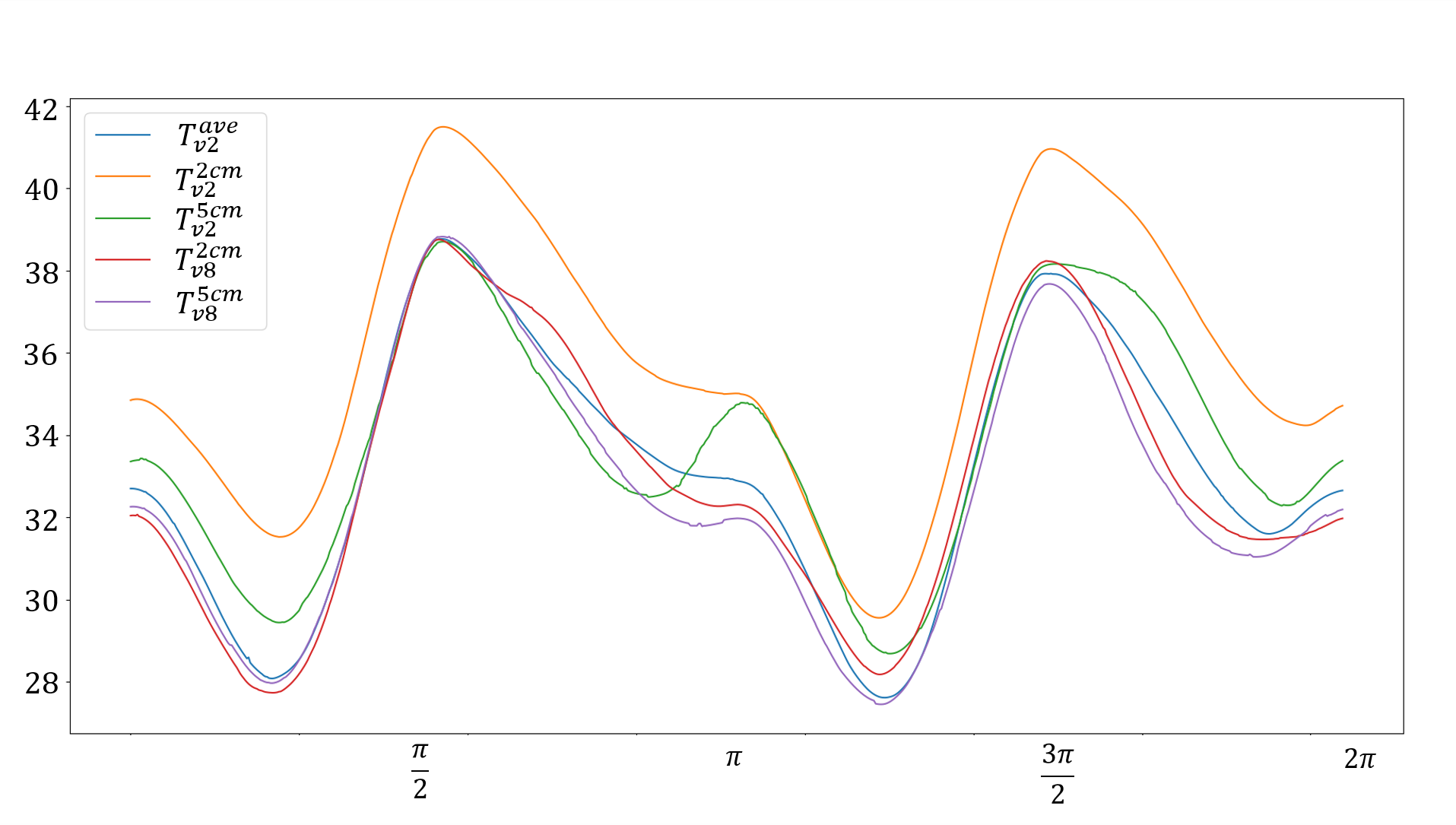}
    \caption{Trajectories of the base height $h_b$ (cm) on different terrains. X-axis denotes the phase angle, from front-left swing [0, 0.5$\pi$] to rear-left swing [1.5$\pi$,2$\pi$]. 
    }
    \label{fig:basePositionCurve}
\end{figure}

\begin{table}[t]
  \scriptsize
  \begin{center}
  \vspace*{0.25cm}
    \caption{Mean and standard deviation of the base height $h_b$, and the target landing height of each foot (cm).}
    \label{tab:basePos}
    \begin{tabular}{r|rrr|rr|rr}
     \hline
     & $T_{v2}^{ave}$ & $T_{v2}^2$ & $T_{v2}^5$ & $T_{v8}^2$ & $T_{v8}^5$ & $T_{c}^2$ & $T_{c}^5$ \\ \hline
    $\sigma (h_b)$ &$3.1$ & $3.2$ & $3.2$ & $3.2$ & $3.1$ & $3.3$ & $2.9$ \\
    $\mu(h_b)$ & $33.1$ & $33.7$ & $32.8$ & $33.0$ & $32.6$ & $35.8$ & $33.5$ \\ \hline
    $\mu(fr)$ & $13.8$ & $13.0$ & $12.6$ & $6.3$ & $6.8$ & $6.9$ & $14.4$ \\
    $\mu(fl)$ & $5.4$ & $5.9$ & $4.5$ & $-0.7$ & $0.2$ & $1.1$ & $6.1$ \\ 
    $\mu(rr)$ & $-1.7$ & $-2.3$ & $-2.9$ & $-7.7$ & $-5.6$ & $-6.3$ & $-1.8$ \\
    $\mu(rl)$ & $-9.7$ & $-10.5$ & $-10.1$ & $-14.8$ & $-11.8$ & $-14.8$ & $-8.9$ \\
    \hline
    \end{tabular}
  \end{center}
\end{table}

\subsubsection{Base height}

Figure \ref{fig:basePositionCurve} shows the trajectories of the base height on different terrains. They show a similar pattern over the terrains: The base descends then ascends during the front leg swing phases ([0,0.5$\pi$] and [$\pi$,1.5$\pi$]) and descends during the rear leg swing phases.
In all scenario, we can observe the smooth movement of the base position.
Table \ref{tab:basePos} shows the means ($\mu$) and standard deviations ($\sigma$) of the base height in various terrain conditions: $T_{v2}^{ave}$ is the value averaged over various terrains in $T_{v2}$, and $T_{v2}^{2}$ is the value on tiles with 2 cm sinking depth in $T_{v2}$. The table shows that the base height has similar means and standard deviations in all scenarios, suggesting that our method produces stable results over terrain variations.  

\subsubsection{Target landing height of each foot}

Table \ref{tab:basePos} also shows the target landing height of each foot ($fl$: front-left, $rr$: rear-right) on different terrains. The height is measured relative to the height of each foot at the default squat pose (Fig. \ref{fig:robot}). 

In all scenarios, the robot learns to take different actions between the left and right legs. Our framework has a fixed phase order starting from the left side, which seems to make the robot rely on the left side more than the right side. 

First, $T_c^2$ and $T_c^5$ show different landing heights for all feet. This suggests that our framework learns different policy depending on the terrain stiffness if the stiffness is constant. Then would it learn to take different actions for terrain with varying stiffness? The target landing heights for $T_{v2}^2$ and $T_{v2}^5$ are similar over all feet, so our framework cannot learn to vary the action against the terrain stiffness if the training environment changes the stiffness every 2 meters. In this scenario, $T_{v2}^{ave}$ is quite similar to $T_c^5$, which suggests that our framework learns to take a conservative policy as if all terrains have the maximum sinking depth if the terrain stiffness changes every 2 meters. 

In contrast,  $T_{v8}^2$ and $T_{v8}^5$ show different actions, especially for the rear feet, showing that our framework learns to take different actions if trained with terrain that changes its stiffness every 8 meters. The terrains $T_c^2$ and $T_{v8}^2$ show very similar results: The robot takes the same action for $T_{v8}^2$ as $T_c^2$. However, $T_{v8}^5$ shows different results from $T_c^5$: The target landing heights of the rear feet of $T_{v8}^5$ are between those of $T_c^2$ and $T_c^5$. This suggests that the framework does not reach the optimal policy for $T_{v8}^5$ terrain. Nonetheless, it is noteworthy that it learns to diverse its action for $T_{v8}^5$ terrain. The standard deviation is less than 1cm for $T_{v8}^2$, but it is around 2cm and 3cm for the frontal and rear feet, respectively, which means that the robot takes much diverse action to step on tiles with different heights on $T_{v8}^5$ terrain.


\subsection{Crucial Components for Non-Rigid Terrain Locomotion}

We discuss key components that make locomotion on non-rigid terrain successful.

\subsubsection{End-effector position history}
The first crucial component is the history of end-effector positions term in observation. It provides the memory of the robot state at four time steps (3 seconds) at the start of each phase, which seems to help the robot cope with the bounciness of the terrain. The learning failed with the memory of even one less phase. We conjecture that providing the end-effector position history also helps the robot adapt its locomotion pattern to the terrain condition; different terrain conditions lead to different end-effector position pattern, which can be used for the robot to take different action strategy.

\subsubsection{End-effector velocity}
The second crucial component of our framework is the end-effector velocity in observation. Without this term, the robot fails to learn to walk, instead it learns just to keep balance without moving forward. 


\subsubsection{Restricted action range}
Another important component of our framework is confining the action space. If the allowed range of action space is too large, the policy fails to learn. When it is too small, the policy either 1) fails by not finding a successful policy to overcome the thresholds made by different tile heights or 2) falls into making inefficient movements, such as moving too little by little. 

\subsection{Other Observation Terms}

\subsubsection{End-effector positions at previous 4 and 8 milliseconds}
The history of the end-effector positions of one cycle is not sufficient to teach the robot whether its foot is currently stuck at a threshold. This additional information helps the robot learn to raise its feet to avoid foot traps or standing still.

\subsubsection{The base orientation terms}
The gravity direction vector and the pitch angle terms help the robot stabilize the base movement, including reducing unnatural  movement such as heading to the sky or to the ground.

\subsection{Adding More Observations}
We conduct several experiments to see whether including more observations obtains a higher reward or reduces the training time. 
First, including a longer history of the end-effector velocities leads to excessive movement of the base. 
Second, including a twice denser history of end-effector positions within one cycle only slows down the learning process significantly. 
Lastly, adding joint position and velocity information does not bring a noticeable change to the result. Since our proposed observation already includes the information on the pose of the robot, adding extra information about joint position and velocity does not seem to benefit.  

\section{Limitations and Future Work}

Our work has a number of limitations that need to be overcome with future research. One major limitation is that our method is not agilely interactive  to the ground because the motion is planned only once per phase while the terrain is dynamically moving when pushed by the feet. This decreases the robot's responsiveness to the change of terrain stiffness and to the case when the foot is caught on the threshold. One way to increase the responsiveness  would be augmenting with an additional lower-level controller that learns to promptly modify the planned motion trajectories according to the terrain conditions. Another more straightforward way would be to design a single-level DRL framework in which the learned policy outputs control commands at each control time step.  

In this work, we only tested with flat elastic terrains. Interesting future work is to explore other types of non-rigid terrains, such as sloped terrains and plastically deforming terrains, which are frequently found in the real world.



\bibliographystyle{IEEEtran}
\bibliography{root.bib}

\end{document}